\documentclass[conference]{IEEEtran}
\IEEEoverridecommandlockouts

\usepackage{cite}
\usepackage{amsmath,amssymb,amsfonts}
\usepackage{graphicx}
\usepackage{textcomp}
\usepackage{xcolor}
\usepackage{booktabs}
\usepackage{multirow}
\usepackage{array}
\usepackage{url}
\usepackage{capt-of}

\graphicspath{{assets/}}
\newcommand{\best}[1]{\textcolor{red}{#1}}
\newcommand{\second}[1]{\textcolor{blue}{#1}}
\newcolumntype{C}[1]{>{\centering\arraybackslash}p{#1}}

\begin{document}

\title{Beyond the Last Layer: Global Inter-Layer Aggregation for AI-Generated Image Detection}

\author{
\IEEEauthorblockN{Wenbin Shen}
\IEEEauthorblockA{
\textit{School of Cyber Science and Engineering}\\
\textit{Nanjing University of Science and Technology}\\
Nanjing, China\\
shenwenbin@njust.edu.cn
}
\and
\IEEEauthorblockN{Zhichao Lian}
\IEEEauthorblockA{
\textit{School of Cyber Science and Engineering}\\
\textit{Nanjing University of Science and Technology}\\
Nanjing, China\\
lzcts@163.com
}
}
\maketitle

\begin{abstract}
The rapid development of generative models has made AI-generated images increasingly close to real images. It has also brought several challenges to detectors, including rapid generator updates, large changes in data distribution, and limited cross-dataset generalization. Many existing methods based on pretrained vision encoders primarily use the global representation from the last layer, potentially leaving useful information from intermediate layers underexplored. To address this issue, we propose Global Inter-Layer Aggregation (GILA). GILA extracts global representations from different depths of a frozen encoder, models interactions among layer representations through inter-layer self-attention, and adaptively aggregates the resulting contextualized representations. The method is not tied to a single pretraining objective and does not require fine-tuning the pretrained encoder. We evaluated GILA on three vision backbones with different pretraining objectives, ten seen generators, and three external datasets. The experimental results show that GILA makes better use of multi-layer information in pretrained encoders and improves detection performance and cross-dataset generalization.
\end{abstract}

\begin{IEEEkeywords}
AI Security, AI-generated Image Detection, Vision Encoder, Inter-layer Interaction, Feature Representation
\end{IEEEkeywords}

\section{Introduction}

The development of generative adversarial networks and diffusion models has greatly lowered the barrier to image generation \cite{b1,b2,b3}. Current generation systems can synthesize images with complex semantics and fine textures. These technologies are useful in content creation, design, and education, but they may also be used for spreading misinformation, identity impersonation, and evading digital forensics. AI-generated image detection has therefore become an important topic in multimedia security \cite{b4}. Early detectors usually learned discriminative cues from spatial artifacts or frequency anomalies left by specific generative models \cite{b5,b6}. However, generator architectures, training data, and post-processing methods continue to change, and detector performance often drops on unseen generators or external datasets. Learning representations that can handle changes in both generative models and data distributions is therefore a central challenge.

Pretrained vision encoders provide a new technical basis for this problem. Existing methods use general representations learned from large-scale natural-image or image--text data and outperform detectors trained from scratch \cite{b7,b12}. However, current detection methods often directly use the global feature from the last encoder layer as the image representation. Research on transfer learning shows that higher network layers gradually become specialized toward the pretraining objective, while feature transferability changes with network depth; intermediate layers can sometimes perform better than higher layers on a target task \cite{b25}. Representation analyses of vision Transformers also show differences in locality, globality, and feature structure across depths \cite{b26}. Several studies further demonstrate that representations from multiple layers contain useful complementary information \cite{b27,b28,b29}. In AI-generated image detection, RINE directly shows the detection value of intermediate encoder-block representations \cite{b13}. These findings indicate that the last layer should not be assumed in advance to be the best choice for every detection setting. This leads to a more basic question: can information interaction across network depths obtain a representation that is more suitable for detection than a single-layer representation?

To address this question, we propose GILA, a global inter-layer aggregation framework for AI-generated image detection. Given a frozen pretrained vision encoder, GILA retains the global representation from each encoding stage and treats network depth as a feature sequence with an explicit order. The inter-layer interaction module allows each layer to update its representation using context from other depths. The adaptive aggregation module then learns layer weights according to the input sample and produces a unified image representation for real/fake classification. GILA does not introduce a new large-scale backbone or rely on an additional frequency prior. Instead, it aims to systematically exploit information that already exists inside a pretrained model but is often ignored when only the last layer is read.

\begin{figure*}[t]
    \centering
    \includegraphics[width=\textwidth]{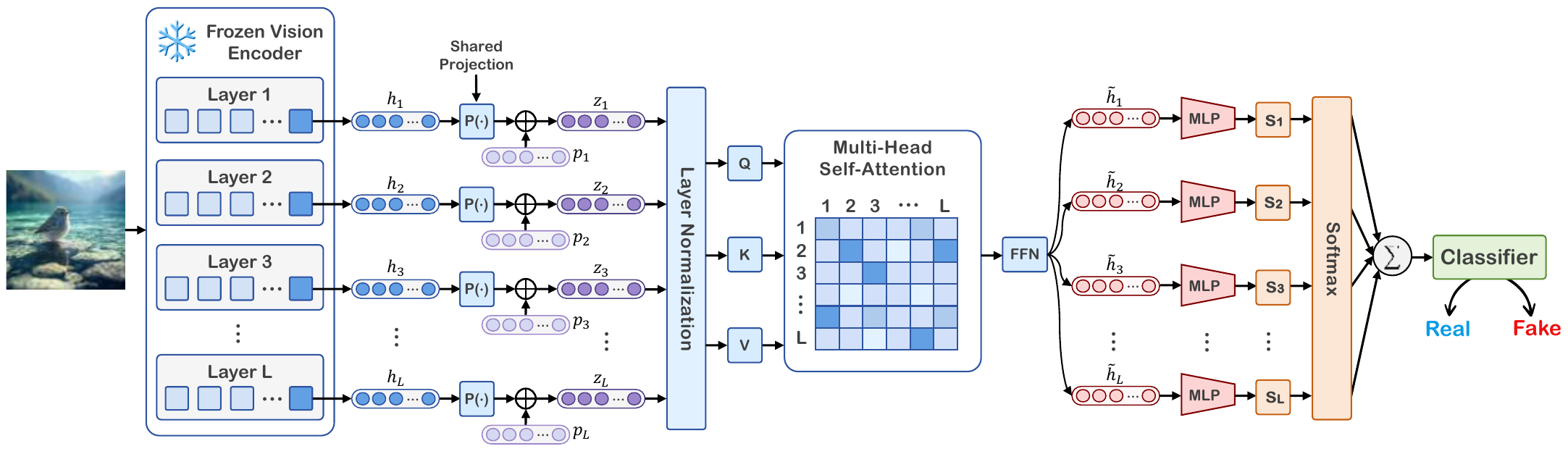}
    \caption{Overall framework of GILA.}
    \label{fig:framework}
\end{figure*}

The main contributions are as follows:
\begin{itemize}
    \item We extend conventional single-layer readout to multi-layer joint modeling, enabling the detector to incorporate hierarchical information across encoder depths.
    \item We propose GILA, a framework combining inter-layer self-attention with sample-adaptive aggregation and applicable to multiple frozen encoders with different pretraining objectives.
    \item We evaluate GILA on three vision backbones, ten seen generators, and three external datasets, and verify its detection performance and cross-dataset generalization.
\end{itemize}

\section{Related Work}

Existing AI-generated image detection methods can be broadly divided into approaches based on local artifacts, frequency cues, reconstruction errors, and pretrained semantic representations. Wang \textit{et al.} found that a convolutional network trained with suitable data augmentation can recognize images generated by different CNN-based models \cite{b5}. Frank \textit{et al.} identified periodic traces left by generator upsampling in the frequency domain \cite{b6}. LGrad uses gradients from a pretrained model to represent local distribution differences in generated images \cite{b8}, while DIRE constructs transferable detection cues from diffusion reconstruction errors \cite{b9}. FreqNet and DFFreq further model generation traces in the frequency domain \cite{b10,b16}. NPR uses neighboring pixel relationships to reduce content bias \cite{b11}. With the development of vision foundation models, UnivFD and later CLIP-based detectors show that frozen or lightly adapted pretrained representations can improve detection of unseen generators \cite{b7,b12}. SAFE preserves local artifacts and reduces overfitting through image transformations and training strategies \cite{b14}, while B-Free points out that biases in training data can lead detectors to learn shortcuts unrelated to the generation mechanism \cite{b15}. These studies have moved detection from specific artifact recognition toward more general representations, but how to make better use of the internal information of pretrained encoders remains open.

Layers at different depths of a deep network do not provide homogeneous representations. Yosinski \textit{et al.} systematically analyzed feature transferability and showed that deep features are affected by specialization to the pretraining task, whereas intermediate features can be more effective for transfer \cite{b25}. Raghu \textit{et al.} further studied the evolution of vision Transformer representations across network depth \cite{b26}. Hypercolumns combine responses from multiple convolutional layers for fine-grained prediction \cite{b27}; FPN fuses features at different scales through a top-down pathway \cite{b29}; and DLA aggregates information across network depths and scales in an iterative and hierarchical manner \cite{b28}. These methods show that the value of multi-layer representations comes not only from simple combination, but also from organizing their relationships and complementary information. In generated-image detection, RINE extracts CLS representations from intermediate CLIP blocks and aggregates them using trainable importance parameters \cite{b13}, demonstrating the detection value of intermediate representations. This method focuses on estimating the relative importance of individual layers, while inter-layer relationships and their adaptation to different samples remain open to further study. Its results are also mainly based on CLIP, and their generality across different pretraining paradigms still needs to be verified.

\section{Methodology}

The overall procedure of GILA consists of multi-layer global feature extraction, layer-wise global interaction, adaptive aggregation, and real/fake classification. The model first reads global representations from different depths of a frozen encoder and projects them into a shared feature space. It then constructs a layer sequence ordered by network depth and uses self-attention to model the relationship between any two layers. Finally, it calculates sample-dependent layer weights from the representations after interaction and uses their weighted sum as the final image feature. Fig.~\ref{fig:framework} shows the overall framework.

\subsection{Feature Extraction}

Let the dataset be
\begin{equation}
    \mathcal{D}=\{(x_i,y_i)\}_{i=1}^{N},
    \label{eq:dataset}
\end{equation}
where $x_i$ denotes the input image, while $y_i=0$ and $y_i=1$ denote a real image and an AI-generated image, respectively. The input image is normalized and resized to match the pretrained encoder, divided into visual units, and then fed into the frozen encoder. Let $T^{(l)}$ denote the output of the $l$-th encoding stage. The model reads only the global representation that summarizes the entire image:
\begin{equation}
    h_l=\mathcal{G}\big(T^{(l)}\big),
    \label{eq:global-feature}
\end{equation}
where $l=1,2,\ldots,L$, and $\mathcal{G}(\cdot)$ denotes the global-token readout operation defined by the encoder. Compared with using all spatial visual units, this representation does not require the construction of a long cross-layer spatial sequence, allowing GILA to focus on the network-depth dimension.

Since the original representation spaces of different encoding stages may not support direct interaction, we use a shared mapping $P$ to project the features from all layers into a common space and add a learnable layer-position embedding $p_l$:
\begin{equation}
    z_l=P(h_l)+p_l.
    \label{eq:projection}
\end{equation}
The features from all layers are stacked in network order:
\begin{equation}
    Z=[z_1,z_2,\ldots,z_L].
    \label{eq:layer-sequence}
\end{equation}
The shared projection provides a common representation space for features from different depths. The layer-position embeddings preserve the network depth of each representation, allowing the interaction module to distinguish different stages and learn the relationship between depth order and detection cues. Thus, $Z$ describes not only a collection of layer features but also their ordered progression through the encoder.

\subsection{Layer-Wise Global Interaction}

If weights are calculated independently for individual layers, each weight can only be estimated from the state of that layer, without determining whether its information is complementary to or redundant with information from other depths. GILA therefore treats $Z$ as a short sequence arranged along network depth and employs a self-attention-based inter-layer interaction module \cite{b30}. For the $m$-th interaction unit, the computation is
\begin{equation}
    A^{(m)}=\operatorname{MSA}\big(\operatorname{LN}(U^{(m-1)})\big),
    \label{eq:msa}
\end{equation}
\begin{equation}
    \bar{U}^{(m)}=U^{(m-1)}+A^{(m)},
    \label{eq:msa-residual}
\end{equation}
\begin{equation}
    U^{(m)}=\bar{U}^{(m)}+\operatorname{FFN}\big(\operatorname{LN}(\bar{U}^{(m)})\big),
    \label{eq:ffn}
\end{equation}
where $U^{(0)}=Z$, and $\operatorname{MSA}$, $\operatorname{FFN}$, and $\operatorname{LN}$ denote multi-head self-attention, a feed-forward network, and layer normalization, respectively. Self-attention is computed along the network-depth dimension, so the updated representation of each layer can refer to all other layers:
\begin{equation}
    \operatorname{Attention}(Q,K,V)
    =\operatorname{softmax}\left(\frac{QK^{\mathsf{T}}}{\sqrt{d}}\right)V.
    \label{eq:attention}
\end{equation}
Here, $Q$, $K$, and $V$ are obtained from the same layer sequence. For each attention head, the attention matrix has a size of $L\times L$, where its $(i,j)$-th element denotes the attention weight assigned to information from the $j$-th depth when updating the representation at the $i$-th depth.
After several interaction units, the contextualized layer representations are obtained as
\begin{equation}
    \widetilde{H}=[\tilde{h}_1,\tilde{h}_2,\ldots,\tilde{h}_L].
    \label{eq:contextualized}
\end{equation}
At this point, $\tilde{h}_l$ is no longer an isolated readout of the $l$-th layer, but a hierarchical representation that contains information from other network depths. Since the sequence length depends only on the number of encoding stages, the computational cost of this module is much lower than applying cross-layer attention to all spatial visual units.

\subsection{Adaptive Aggregation and Classification}

Different images may rely on detection evidence from different depths. Therefore, for the representation of the $l$-th layer after interaction, GILA uses a lightweight scoring function to calculate its importance:
\begin{equation}
    s_l=w_2^{\mathsf{T}}\phi\big(W_1\operatorname{LN}(\tilde{h}_l)\big),
    \label{eq:score}
\end{equation}
where $W_1$ and $w_2$ are learnable parameters, and $\phi(\cdot)$ denotes the GELU activation function. The scores are then normalized over all layers:
\begin{equation}
    \alpha_l=\frac{\exp(s_l)}{\sum_{j=1}^{L}\exp(s_j)}.
    \label{eq:alpha}
\end{equation}
The normalized weights are non-negative and sum to one. Because they are calculated from the contextualized representations of the current image, the resulting layer distribution can vary across samples instead of remaining fixed for the entire dataset.
Accordingly, the final image representation is
\begin{equation}
    v=\sum_{l=1}^{L}\alpha_l\tilde{h}_l.
    \label{eq:pooled}
\end{equation}
This weighted aggregation avoids specifying an optimal network layer in advance and instead retains detection evidence from different depths according to the input image. Next, the aggregated representation $v$ undergoes layer normalization and a fully connected transformation, followed by GELU activation to obtain a compact hidden representation:
\begin{equation}
    r=\phi\!\left(W_h\operatorname{LN}(v)+b_h\right),
    \label{eq:hidden-classifier}
\end{equation}
where $W_h$ and $b_h$ are learnable parameters, $\phi(\cdot)$ denotes the GELU activation function, and $\operatorname{LN}(\cdot)$ denotes layer normalization. The output layer then produces prediction logits for the real and generated classes, which are finally mapped to prediction probabilities:
\begin{equation}
    p=\operatorname{softmax}\big(W_o\mathcal{R}_{\rho}(r)+b_o\big),
    \label{eq:classifier}
\end{equation}
where $W_o$ and $b_o$ are learnable parameters, and $\mathcal{R}_{\rho}(\cdot)$ denotes the dropout operation with dropout rate $\rho$. The model uses cross-entropy loss to jointly optimize its trainable components:
\begin{equation}
    \mathcal{L}_{\mathrm{cls}}
    =-\frac{1}{N}\sum_{i=1}^{N}\sum_{c\in\{0,1\}}
    \mathbb{I}(y_i=c)\log p_{i,c}.
    \label{eq:loss}
\end{equation}
The classification loss provides a unified supervisory objective for inter-layer interaction, layer-importance estimation, and final prediction. In this way, GILA forms an integrated process from multi-layer global feature extraction and cross-layer contextualization to sample-adaptive aggregation, allowing representations at different depths to contribute jointly to real-versus-generated discrimination.

\begin{table*}[t]
\caption{Seen-generator detection accuracy (\%). Best and second-best results are shown in red and blue, respectively.}
\label{tab:seen}
\centering
\setlength{\tabcolsep}{2.5pt}
\begin{tabular}{l*{11}{C{0.49in}}}
\toprule
Method & DALL-E & Firefly & Midjourney & SDXL & GPT-4o & Kling & MiniMax & FLUX & Jimeng & Nano & \textbf{Overall} \\
\midrule
LGrad  & 84.95 & 72.33 & 74.03 & 79.61 & 81.73 & 85.38 & 72.14 & 75.88 & 77.78 & 75.75 & \textbf{77.93} \\
FreqNet & 76.70 & 70.87 & 79.37 & 79.37 & 82.74 & 80.77 & 78.86 & 75.38 & 82.58 & 82.75 & \textbf{78.90} \\
SAFE & 88.35 & 92.72 & 80.58 & 92.96 & 93.65 & 81.03 & 87.56 & 85.93 & 91.67 & 68.00 & \textbf{86.27} \\
UnivFD & 92.72 & 89.56 & 90.29 & 92.72 & 92.64 & 91.54 & 85.82 & 85.43 & 92.42 & 92.75 & \textbf{90.59} \\
DFFreq & 94.66 & 90.53 & 88.83 & 91.99 & 93.15 & 90.77 & 81.84 & 89.20 & 93.69 & 93.50 & \textbf{90.81} \\
NPR & 92.72 & 83.01 & 91.26 & 94.17 & 95.43 & 93.08 & 84.83 & 90.20 & 96.21 & 95.00 & \textbf{91.56} \\
RINE & \best{100.00} & 87.14 & 95.15 & 97.33 & 96.45 & 96.41 & 91.29 & 98.49 & \second{99.24} & \second{99.25} & \textbf{96.05} \\
\midrule
GILA-M (Ours) & 97.33 & 94.42 & 95.87 & 97.57 & \second{98.98} & \second{98.97} & 95.27 & \second{98.74} & 98.99 & 97.50 & \textbf{97.34} \\
GILA-D (Ours) & 98.54 & \second{94.66} & \second{96.60} & \second{98.30} & 98.73 & 97.44 & \second{96.52} & 97.74 & \second{99.24} & 98.75 & \textbf{\second{97.64}} \\
GILA-C (Ours) & \second{99.27} & \best{98.06} & \best{99.03} & \best{99.27} & \best{99.75} & \best{99.74} & \best{100.00} & \best{100.00} & \best{100.00} & \best{100.00} & \textbf{\best{99.50}} \\
\bottomrule
\end{tabular}
\end{table*}

\section{Experiments}

\subsection{Experimental Settings}

\textbf{Datasets.} This study uses five datasets: Synthbuster \cite{b17}, AIGI-Now \cite{b18}, FakeBench \cite{b19}, SynthWildX \cite{b12}, and WildRF \cite{b20}. Seen-generator detection uses Synthbuster and AIGI-Now. Synthbuster contains DALL-E, Firefly, Midjourney, and SDXL, while AIGI-Now contains GPT-4o, Kling, MiniMax, FLUX, Jimeng, and Nano, covering ten generators in total. The two datasets are divided into training, validation, and test sets with an approximate ratio of $3:1:1$. During the split, a generated image and its corresponding real reference are treated as associated samples. The same real source and all of its associated generated images are kept in the same subset to avoid content leakage. Models trained on these two datasets are directly tested on FakeBench, SynthWildX, and WildRF to evaluate generalization across different image sources, generative models, and data distributions.

\textbf{Baselines.} We select representative baselines covering spatial, gradient, frequency, reconstruction, and pretrained-representation approaches. The compared methods include LGrad \cite{b8}, UnivFD \cite{b7}, FreqNet \cite{b10}, NPR \cite{b11}, SAFE \cite{b14}, DFFreq \cite{b16}, and RINE \cite{b13}. For a fair comparison, all baselines use the same datasets and data splits as our method. Shared training settings are kept consistent where applicable, while method-specific hyperparameters follow the corresponding original papers.

\begin{table}[t]
\caption{Ablation results for different components of GILA (\%). Best and second-best results within each backbone group are shown in red and blue, respectively. ILI denotes inter-layer interaction, and AA denotes adaptive aggregation.}
\label{tab:ablation}
\centering
\setlength{\tabcolsep}{3.5pt}
\begin{tabular}{l*{5}{C{0.42in}}}
\toprule
Method & Acc. & Pre. & Rec. & F1 & AUC \\
\midrule
Last-Layer-M & 96.13 & \second{96.55} & 95.67 & 96.11 & 99.28 \\
\hspace{0.35em}w/o ILI-M & 92.35 & 90.77 & 94.29 & 92.50 & 97.93 \\
\hspace{0.35em}w/o AA-M & \second{96.43} & 96.38 & \second{96.47} & \second{96.43} & \second{99.43} \\
\textbf{GILA-M} & \textbf{\best{97.34}} & \textbf{\best{97.13}} & \textbf{\best{97.57}} & \textbf{\best{97.35}} & \textbf{\best{99.51}} \\
\midrule
Last-Layer-D & 88.85 & 88.04 & 89.92 & 88.97 & 95.39 \\
\hspace{0.35em}w/o ILI-D & 95.13 & 92.92 & \best{97.72} & 95.26 & \second{99.24} \\
\hspace{0.35em}w/o AA-D & \second{95.43} & \second{93.53} & \second{97.62} & \second{95.53} & 98.90 \\
\textbf{GILA-D} & \textbf{\best{97.64}} & \textbf{\best{98.05}} & \textbf{97.22} & \textbf{\best{97.63}} & \textbf{\best{99.35}} \\
\midrule
Last-Layer-C & 96.85 & 96.03 & 97.74 & 96.88 & 99.64 \\
\hspace{0.35em}w/o ILI-C & \second{99.01} & \second{99.25} & 98.76 & \second{99.00} & \best{99.91} \\
\hspace{0.35em}w/o AA-C & 98.31 & 97.28 & \second{99.40} & 98.33 & \second{99.86} \\
\textbf{GILA-C} & \textbf{\best{99.50}} & \textbf{\best{99.41}} & \textbf{\best{99.60}} & \textbf{\best{99.50}} & \textbf{99.79} \\
\bottomrule
\end{tabular}
\end{table}

\textbf{Implementation Details.} To evaluate the applicability of the method across different pretraining objectives and backbone types, we use three frozen vision encoders: DINOv2-base \cite{b24}, CLIP ViT-B/16 \cite{b22}, and ViT-MAE-base \cite{b23}. They are denoted by GILA-D, GILA-C, and GILA-M, respectively. For each encoder, we extract 768-dimensional global representations from 12 stages, and use an input size of $448\times448$. GILA projects these features into a 512-dimensional shared space. The inter-layer interaction module contains two Transformer encoder blocks. Each block uses eight attention heads, a feed-forward expansion ratio of four, and a dropout rate of 0.2. The classification head consists of layer normalization, a fully connected layer, GELU, dropout, and a two-class output layer. The model is optimized with AdamW \cite{b31}. The initial learning rate and weight decay are both $10^{-4}$, and the batch size is 64. Each model is trained for at most 50 epochs. Training stops early if the validation metric does not improve for five consecutive epochs.

\subsection{Seen-Generator Detection}
Table~\ref{tab:seen} reports the detection accuracy of different methods on ten seen generators. All three GILA configurations outperform the evaluated baselines, with overall accuracies of 97.34\%, 97.64\%, and 99.50\%, respectively. GILA-C achieves the highest accuracy on nine of the ten generators, with DALL-E as the only exception. Compared with the best-performing baseline, RINE (96.05\%), GILA-C improves the overall accuracy by 3.45 percentage points. Even GILA-M, which has the lowest overall accuracy among the three configurations, still outperforms the strongest baseline. These results suggest that the observed benefits of inter-layer aggregation are not limited to one of the three evaluated pretrained encoders.

\begin{table*}[t]
\caption{Cross-dataset generalization results (\%). Best and second-best results are shown in red and blue, respectively.}
\label{tab:cross-dataset}
\centering
\setlength{\tabcolsep}{2.5pt}
\begin{tabular}{l*{9}{C{0.52in}}}
\toprule
\multirow{2}{*}{Method} & \multicolumn{3}{c}{FakeBench} & \multicolumn{3}{c}{SynthWildX} & \multicolumn{3}{c}{WildRF} \\
\cmidrule(lr){2-4}\cmidrule(lr){5-7}\cmidrule(lr){8-10}
& Acc. & F1 & AUC & Acc. & F1 & AUC & Acc. & F1 & AUC \\
\midrule
LGrad & 63.63 & 56.10 & 68.49 & 56.93 & 64.65 & 68.15 & 61.36 & 64.51 & 65.70 \\
FreqNet & 70.98 & 71.10 & 75.18 & 69.18 & 79.99 & 58.99 & 49.49 & 57.95 & 46.14 \\
SAFE & 71.07 & 60.20 & 89.55 & 70.90 & 81.55 & 58.03 & 44.85 & 55.92 & 40.01 \\
UnivFD & 68.07 & 61.34 & 74.01 & 80.56 & 88.51 & 63.28 & 53.77 & 65.19 & 52.49 \\
DFFreq & 74.58 & 69.62 & 81.79 & 76.48 & 84.49 & 74.23 & 58.01 & 61.45 & 60.81 \\
NPR & 75.65 & 71.01 & 80.87 & 74.41 & 83.03 & 72.63 & 58.37 & 61.78 & 61.60 \\
RINE & 68.42 & 54.26 & 89.74 & 82.35 & 89.45 & \best{90.85} & 62.83 & 70.00 & \second{76.45} \\
\midrule
GILA-M (Ours) & \best{77.02} & \best{74.61} & 87.22 & \best{83.61} & \best{89.79} & \second{87.15} & \second{66.85} & \second{72.49} & 68.41 \\
GILA-D (Ours) & 73.27 & 63.79 & \best{96.23} & 82.40 & 88.85 & 81.38 & \best{72.56} & \best{76.00} & \best{81.33} \\
GILA-C (Ours) & \second{76.93} & \second{71.18} & \second{93.87} & \second{83.44} & \second{89.58} & 86.28 & 63.32 & 66.96 & 67.99 \\
\bottomrule
\end{tabular}
\end{table*}

\begin{figure*}[!t]
    \centering
    \begin{minipage}[t]{0.48\textwidth}
        \centering
        \includegraphics[width=\linewidth]{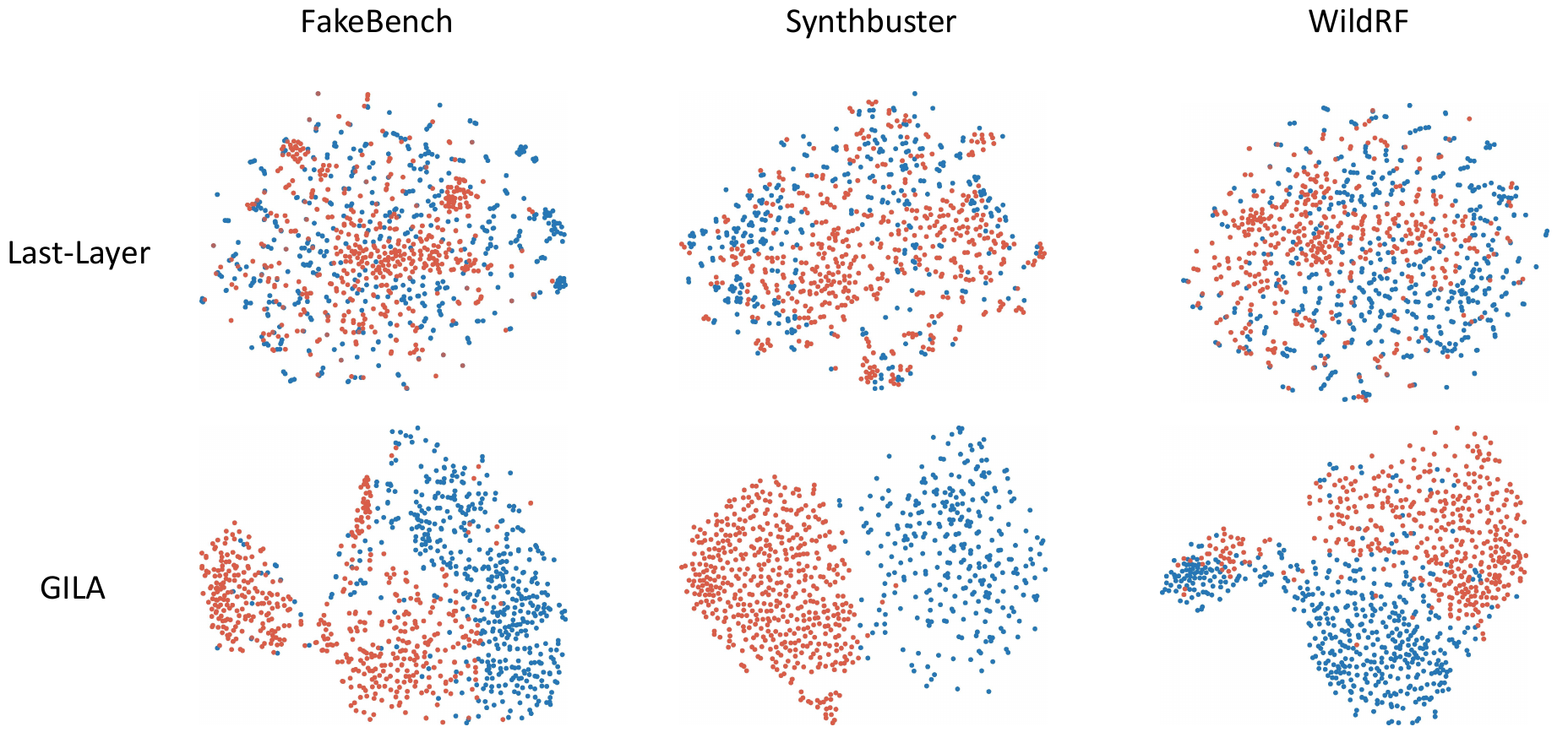}
        \captionof{figure}{t-SNE visualization of last-layer features and GILA representations.}
        \label{fig:tsne}
    \end{minipage}
    \hfill
    \begin{minipage}[t]{0.48\textwidth}
        \centering
        \includegraphics[width=\linewidth]{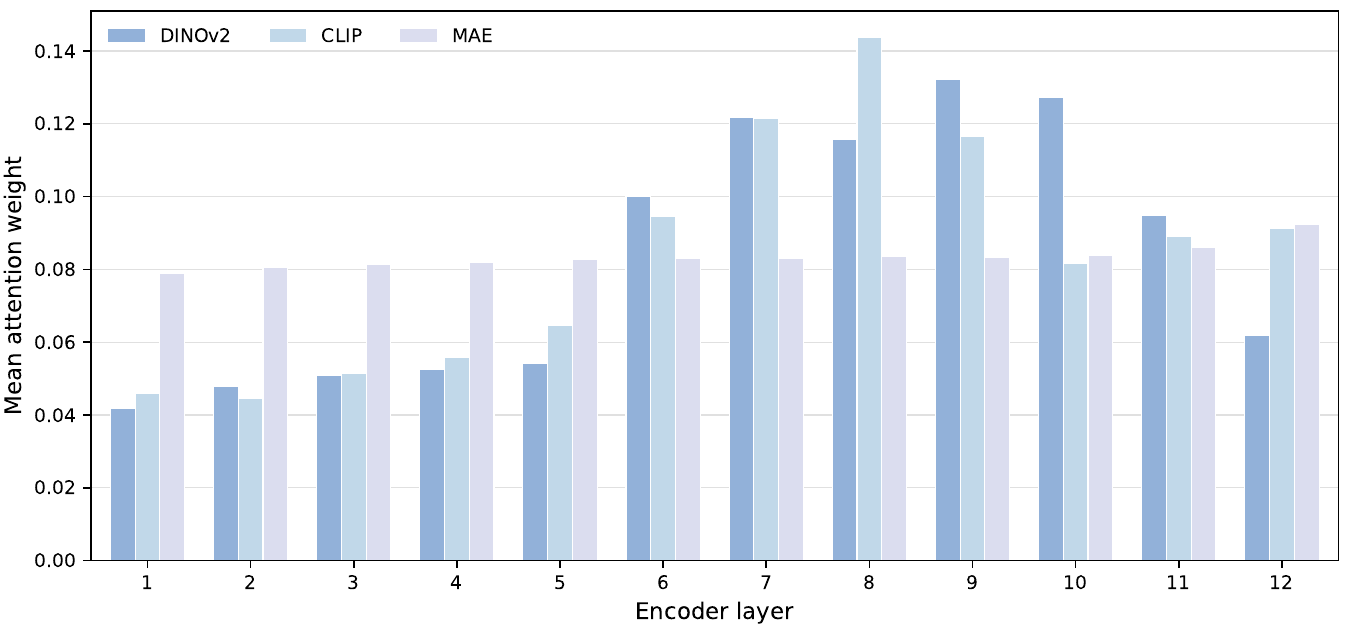}
        \captionof{figure}{Average layer-attention distributions for the three vision backbones.}
        \label{fig:layer-attention}
    \end{minipage}
\end{figure*}

\subsection{Ablation Study}
Table~\ref{tab:ablation} evaluates GILA in terms of multi-layer representations, inter-layer interaction, and adaptive aggregation. Compared with last-layer readout, the complete GILA improves accuracy by 1.21, 8.79, and 2.65 percentage points on MAE, DINOv2, and CLIP, respectively. Removing inter-layer interaction or replacing adaptive aggregation with uniform aggregation consistently reduces accuracy, with drops of up to 4.99 and 2.21 percentage points, respectively. These results suggest that intermediate-layer information is useful and that both modules contribute to the overall performance.

\subsection{Cross-Dataset Generalization}

Table~\ref{tab:cross-dataset} reports the cross-dataset generalization results. Compared with the existing baseline that achieves the highest accuracy on each target dataset, GILA improves accuracy by 1.37, 1.26, and 9.73 percentage points on FakeBench, SynthWildX, and WildRF, respectively. On FakeBench, GILA-M achieves the highest accuracy and F1, while GILA-D achieves the highest AUC. On SynthWildX, GILA-M achieves the highest accuracy and F1. On WildRF, GILA-D achieves the best results on all three metrics. These results show that different pretrained backbones have different levels of adaptability to the target data distributions. MAE performs better on FakeBench and SynthWildX, whereas DINOv2 shows stronger generalization on WildRF.

\subsection{Visualization Analysis}

Fig.~\ref{fig:tsne} uses t-SNE \cite{b32} to project last-layer features and GILA representations into two dimensions. Using CLIP as an example, real and generated samples overlap over a large area in the last-layer feature spaces of FakeBench, Synthbuster, and WildRF. After inter-layer interaction and adaptive aggregation, the two classes form clearer clusters with more distinct boundaries. Qualitatively, the t-SNE projections show greater visual separation between real and generated samples after applying GILA. This observation is consistent with the quantitative improvements reported in Tables~\ref{tab:seen} and~\ref{tab:cross-dataset}.

Fig.~\ref{fig:layer-attention} shows the average layer weights of the three backbones over all test data. The weight distributions of DINOv2 and CLIP are clearly non-uniform. Their higher weights mainly appear in the middle and later layers rather than being concentrated only in the last layer. The MAE distribution is relatively flat, although small changes remain in the last few layers. The observed weighting patterns differ across the three pretrained encoders, suggesting that GILA uses information from different depths differently depending on the encoder.

\section{Conclusion}

This paper studies the underuse of multi-layer global representations from pretrained vision encoders in AI-generated image detection and proposes GILA. The method organizes global features from different network depths as an ordered sequence, models inter-layer interactions, and adaptively aggregates the resulting representations for real/fake classification. Results with three pretrained backbones show that GILA improves over last-layer readout. Experiments on ten seen generators and three external datasets demonstrate performance gains on both seen-generator and cross-dataset evaluations under the considered settings.

\bibliographystyle{IEEEtran}
\bibliography{references}

\end{document}